\documentclass[journal]{IEEEtran}
\usepackage{amsmath,amssymb}
\usepackage{psfrag}
\usepackage{epsfig}
\usepackage{cite}
\usepackage{graphics}
\usepackage{color}
\usepackage{subfigure}
\usepackage{multirow}
\usepackage{threeparttable}
\usepackage{booktabs}
\usepackage{mathrsfs}
\usepackage{bm}
\usepackage[ruled,vlined]{algorithm2e}

\usepackage{hyperref}
\usepackage{breakurl}

\begin{document}
\title{Beetle Antennae Search without Parameter Tuning (BAS-WPT) for Multi-objective Optimization}
\author{Xiangyuan Jiang, Shuai Li
%
}
\markboth{}%
{Shell \MakeLowercase{\textit{et al.}}: Bare Demo of IEEEtran.cls
for Journals} \maketitle
\begin{abstract}
Beetle antennae search (BAS) is an efficient meta-heuristic algorithm inspired by foraging behaviors of beetles. This algorithm includes several parameters for tuning and the existing results are limited to solve single objective optimization. This work pushes forward the research on BAS by providing one variant that releases the tuning parameters and is able to handle multi-objective optimization. This new approach applies normalization to simplify the original algorithm and uses a penalty function to exploit infeasible solutions with low constraint violation to solve the constraint optimization problem. Extensive experimental studies are carried out and the results reveal efficacy of the proposed approach to constraint handling.
\end{abstract}


\section{Introduction}

Nature-inspired algorithms have giant potential to solve optimization problems and have been successfully implemented in various scientific and engineering domains \cite{gaiso}. main challenges of meta-heuristic algorithms lie in how to handle various constraints imposed on variables and how to simplify the parameter tuning of algorithms.

Because of the presence of constraints, the feasible space may be largely reduced, making the searching a changeling task a comparison with non-constrained optimization with a single objective. To solve the aforementioned optimization problem, Tessema and Yen \cite{aapff} propose an adaptive penalty function to exploit infeasible solutions with appropriate fitness value and low constraint violation. And then, they extend the constraint-handling results to multi-objective evolutionary optimization problem based on adaptive penalty function and distance measure \cite{chime}. Using a multi-objective formulation, Runarsson and Yao \cite{sbice} propose a common approach to apply a penalty function to bias the search towards a feasible solution.

In terms of parameter tuning, we further develop our previous work call the beetle antennae search (BAS) algorithm which is inspired by the searching and detecting behavior of longhorn beetles. In this paper, we improve the result in \cite{BAS} to be simple to implement and need no parameter tuning. As the original BAS in \cite{BAS} does not consider constraint, we also modify the original BAS to be capable of solving constrained optimization problem simultaneously. The main contribution of the are stated as below:
\begin{itemize}
  \item{1} Normalization method is used to extend the original BAS algorithm to general form without parameter tuning, which makes the algorithm implemented simply.
  \item{2} Based on the improved BAS algorithm without parameter tuning (BAS-WPT), constraint optimization problems are formulated a multi-objective optimization problem and further handled by penalty function method.
\end{itemize}

The rest of the paper is organized as follows. In Section \ref{sec.BAS-WPT}, improved BAS-WPT algorithm is proposed. In Section \ref{sec.constraint}, BAS-WPT are used in constraint optimization problem. In Section \ref{sec.experiment}, numerical results are presented and compared. In Section \ref{sec.conclusion}, a concluding remark is drawn.

\section{Proposed Approach Design}\label{sec.BAS-WPT}

In this section, by considering the original BAS, an improved BAS is presented without parameter tuning to simply the application for user. The algorithm is basically an original implementation except for the normalization of input data.

\subsection{the Original BAS}

For clear illustration, the original BAS is included in Algorithm \ref{alg.BAS-WPT}, which is capable of searching global optimum of both convex and non-convex problem in a general function:

\begin{equation*}
  \frac{\text{Minimize}}{\text{Maximize}} f(\bm{x}), \bm{x}=[x_1,x_2,\cdots,x_N]^\text{T}
\end{equation*}
where $f$ is the fitness function and $\bm{x} \in \mathbb{R}^N$ denotes the input data in $n$ dimensions.
The main formula of the natural-inspired BAS consist of tow aspect: searching behavior and detecting behavior. The searching behavior is used to explore by introducing a normalized random unit vector $\overrightarrow{\bm{b}}$ to enhance the searching ability,
\begin{eqnarray}\label{eqn.search}
    \bm{x}_r&=&\bm{x}^t+d^t\overrightarrow{\bm{b}}, \nonumber\\
  \bm{x}_l&=&\bm{x}^t-d^t\overrightarrow{\bm{b}},
\end{eqnarray}
and the detecting behavior is used to exploit in an iterative form,
\begin{equation}\label{eqn.detect}
  \bm{x}^t=\bm{x}^{t-1}+\delta^t\overrightarrow{\bm{b}}\text{sign}(f(\bm{x}_r)-f(\bm{x}_l)),
\end{equation}
where $d$ represents the distance between tow antennae of a longhorn beetle and $\delta$ represents the step size of each iteration.

Evidently, the presetting of parameters such as $d$ and $\delta$ influences performance of BSA seriously. Thus, we attempt to develop a much effective and robust improvement.

\subsection{the Proposed BAS-WPT Approach}
\begin{figure}[t]\centering
\subfigure[]{\includegraphics[scale=0.28]{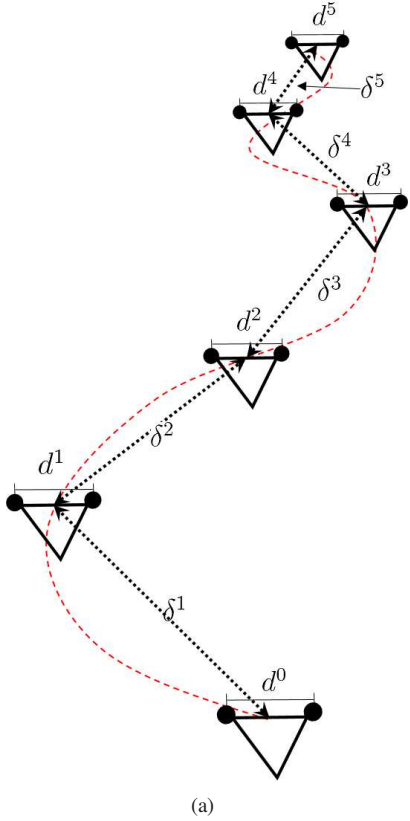}}
\caption{The iterative optimization process of BAS-WPT in $5$ steps. The dotted line in red color denotes the trajectory of fitness function. The triangle represents a beetle, on both sides of which two solid circles denote antennae of the beetle, $d$ is the distance between tow antennae and $\delta$, corresponding to the dotted line in black color, is the step size of searching.}\label{fig.BAS}
\end{figure}

Fig. \ref{fig.BAS} demonstrates the iterative optimization precess of BAS-WPT which can be seen as a variable scale algorithm from the figure obviously.

For the sake of simplicity, we use normalization method to tune parameters of BAS adaptively. Assume that $x_i$, the $i$th element of $\bm{x}$, lies in the rang from $x_{i,l}$ to $x_{i,h}$, and then $\bm{x}$ satisfied $\bm{x} \in [\bm{x}_l,\bm{x}_h]$, where $\bm{x}_l$ is lower bound and $\bm{x}_h$ is the upper bound. The input data used in fitness functon could be formulated in the following expression at each iteration:
\begin{equation}\label{eqn.inversenorm}
  \tilde{x}_i=x_i( x^\text{max}_i - x^\text{min}_i )+ x^\text{min}_i.
\end{equation}
Finally, we obtain the global optimum $f(\tilde{\bm{x}})_\text{bst}$ corresponding to the position $\tilde{\bm{x}}_\text{bst}$ by BAS-WPT algorithm with normalized variant $\bm{x}_\text{bst}$.

To simplify the parameter tuning further more, we also construct the relationship between searching distance $d$ and step size $\delta$ as follows:
\begin{eqnarray}\label{eqn.parameter}
  \delta^t&=&c_1\delta^{t-1}+\delta^0,
  d^t=\delta^t/c_2,
\end{eqnarray}
where $c_1$ and $c_2$ are constants to be adjusted by designers.

\section{Constraint Handling by BAS-WPT}\label{sec.constraint}

In this section, we extend the BAS-WPT algorithm into constrained optimization problem with penalty function method.

\subsection{Problem Formulation}

A constraint optimization problem can be formulated as
\begin{eqnarray}
  &&\frac{\text{Minimize}}{\text{Maximize}} f(\bm{x}),\nonumber\\
  &s.t. &g_j(\bm{x})\leq 0, j=1, \cdots, K, \nonumber\\
        &&x^\text{max}_i \leq x_i \leq x^\text{min}_i, i=1, \cdots, N,
\end{eqnarray}

There are $K$ inequality function constraints required to be satisfied by the optimal solution. The presence of constraints of both inequality functions and variants restrict the searching area to be a interest region, where suitable solution could be found.
\subsection{Penalty Function Method}

To solve the constrained optimization problem, we present penalty function method to deal with inequality function constraint.

In penalty functions, infeasible solutions are penalized for the violation of the inequality constraint by putting penalty terms on the original fitness function, which will reduce the probability of selecting an infeasible solution. Specially, penalty function in our study is formulated in the following form:
\begin{equation}\label{equ.penalty}
  F(\bm{x})=f(\bm{x})+\lambda\sum_{j=1}^{K}h_j(\bm{x})g_j(\bm{x}),
\end{equation}
where $F(\bm{x})$ is the improved fitness function, $f(\bm{x})$ is the original fitness function, $\lambda$ is the penalty parameter usually predefined as a large enough value (e.g. $10^10$), and the constraint violation $h_j(\bm{x})$ is defined as
\begin{equation}\label{eqn.cons}
  h_j(\bm{x})=\left\{ \begin{array}{ll}
    1, & g_j(\bm{x})>0\\
    0, & g_j(\bm{x})\leq0
  \end{array}
   \right.
\end{equation}
When anyone of the inequality constraints $g_j(\bm{x})>0$ satisfies accompanying with a large value $\lambda$, the second term of (\ref{equ.penalty}) dominates the fitness function, which makes $F(\bm{x}) \to \infty$. Otherwise,  all $h_j(\bm{x})=0$ are satisfied, and thus $F(\bm{x})=f(\bm{x})$.

Algorithm \ref{alg.BAS-WPT} corresponding to BAS-WPT algorithm demonstrates the improvement based on BAS adopted in the research to design a more feasible approach to solve the constraint optimization problem.
\begin{algorithm}\label{alg.BAS-WPT}
\caption{BAS-WPT algorithm for constrained optimization}

\KwIn{Initialize the input data $\bm{x}^0$ at $0$ time instance in standard normalization form
$x^0_i=(\text{rnd}(\cdot)-x^\text{min}_i)/(x^\text{max}_i - x^\text{min}_i)$ for each element
,  and initialize the parameters $c_1, c_2, \delta^0$.}
\KwOut{$\bm{x}_\text{bst}$, $f_\text{bst}$.}
\While{ ($t<T_\text{max}$) or (stop criterion)}{
Search in variable space with two kinds of antennae according to (\ref{eqn.search})\;
Update the state variable $\bm{x}^t$ according to (\ref{eqn.detect})\;
Generate the normalized vector $\tilde{\bm{x}}$ according to (\ref{eqn.parameter})\;
Construct the improved fitness function according to (\ref{equ.penalty}) and (\ref{eqn.cons})\;
\If{$F(\bm{x}^t)$ satisfies optimum condition}
{$F_\text{bst}=F(\bm{x}^t)$, $\bm{x}_\text{bst}=\bm{x}^t$.}
Update parameters according to (\ref{eqn.parameter}).
Calculate the best potion $\tilde{\bm{x}}_\text{bst}$ by $\bm{x}_\text{bst}$ similarly to (\ref{eqn.inversenorm}).
}

\Return{$\bm{x}_\text{bst}$, $f_\text{bst}$.}
\end{algorithm}

\section{Experimental Study}\label{sec.experiment}

In this section, we present tow examples from optimization literatures to demonstrate the performance of the proposed BAS-WPT algorithm for constrained optimization.

\subsection{Pressure Vessel Function}

There are four variables in pressure vessel problem which aims at minimizing the fitness function below:
\begin{eqnarray*}
  \text{minimize} f(\bm{x}) &=&0.6224x_1x_3x_4+1.7781x_2x^2_3 \nonumber\\
&&+3.1661x^2_1x_4 + 19.84x^2_1x_3,\nonumber\\
s.t. ~~ g_1(\bm{x}) &=& -x1 + 0.0193x_3 \leq 0,\nonumber\\
 g_2(\bm{x}) &=& -x_2 + 0.00954x_3 \leq 0,\nonumber\\
 g_3(\bm{x}) &=& -\pi x^2_3x_4 -\frac{4}{3}\pi x^3_3 + 1296000 \leq 0,\nonumber\\
 g_4(\bm{x}) &=& x_4-240\leq 0,\nonumber\\
 x_1 &\in& \{1,2,3,\cdots,99\}\times0.0625,\\
 x_2 &\in& \{1,2,3,\cdots,99\}\times0.0625,\\
 x_3 &\in& [10,200],\\
 x_4 &\in& [10,200].
\end{eqnarray*}

Table \ref{table1} illustrates the best results obtained by the proposed BAS-WPT algorithm using only $150$ iterations and other various existing algorithms to solve the pressure vessel optimization problem. It is worth pointing out that the best result from the proposed BAS-WPT algorithm is better than most of the existing ones and has the fastest convergence simultaneously.

\begin{table*}[!t]
  \footnotesize
  \centering
  \caption{Comparisons of results for Pressure Vessel Function}
  \begin{tabular} {c c c c c c c c c c}
  \hline
&$x_1$&$x_2$&$ x_3$&$x_4$&$g_1({\bm{x}})$&$g_2({\bm{x}})$&$g_3({\bm{x}})$&$g_4({\bm{x}})$ & $f_*$\\
\hline
  \cite{aiaco} & 0.8125&0.4375&42.0984 &176.6378& -8.8000e-7& -0.0359&-3.5586&-63.3622& 6059.7258 \\
    \cite{agafn}& 1 &0.625&   51.2519 &   90.9913  &   -1.011  &   -0.136  &   -18759.75  &  -149.009  & 7172.300   \\
   \cite{hagaw}&0.8125& 0.4375  & 42.0870 & 176.7791& -2.210e-4 &-0.03599&  -3.51084   &  -63.2208  &  6061.1229  \\
            \cite{gofsd}     &1.000& 0.625& 51.000& 91.000 &-0.0157& -0.1385 &-3233.916& -149 &7079.037\\
            \cite{assmf}     &0.8125 &0.4375& 41.9768& 182.2845 &-0.0023& -0.0370& -22888.07 &-57.7155& 6171.000\\
             \cite{aalmb}    &1.125& 0.625 &58.291& 43.690 &0.000016& -0.0689 &-21.2201& -196.3100 &7198.0428\\
             \cite{garod}    &0.9375 &0.5000& 48.3290 &112.6790& -0.0048& -0.0389 &-3652.877& -127.3210& 6410.3811\\
             \cite{uoasp}    &0.8125 &0.4375 &40.3239& 200.0000 &-0.034324& -0.05285& -27.10585& -40.0000 &6288.7445\\
             \cite{gafnm}    &1.125& 0.625& 58.1978& 44.2930 &-0.00178& -0.06979& -974.3 &-195.707& 7207.494\\
             \cite{niadp}    &1.125& 0.625& 48.97& 106.72& -0.1799& -0.1578& 97.760 &-132.28& 7980.894\\
             \cite{anmaf}    &1.125& 0.625 &58.2789 &43.7549 &-0.0002& -0.06902& -3.71629& -196.245 &7198.433\\
             \cite{fastf}    &0.7782& 0.3846& 40.3196& 200.000& -3.172e-5& 4.8984e-5$\diamond$&1.3312$\diamond$&-40& 5885.33\\
  BAS-WPT &0.8125&0.4375&42.09355&42.09355&-9.43E-05&-0.03592&-413.6252&-63.2285&6062.04676\\
\hline
\end{tabular} \label{table1}
\begin{tablenotes}
 \item $\diamond$ denotes the violation of the corresponding constraint.
\end{tablenotes}
\end{table*}

\subsection{Himmelblau Function}

We also consider the Himmelblau' nonlinear optimization problem which is a famous benchmark used used for several evolutionary algorithm before. The problem consists of 5 variables, 6 inequality constraint and 10 boundary conditions and could be further stated as follows:

\begin{eqnarray*}
  \text{minimize} f(\bm{x}) &=& 5.3578547x^2_3 +0.8356891x_1x_5 \nonumber\\
&&+ 37.29329x_1 - 40792.141,\nonumber\\
s.t. ~~g_1(\bm{x}) &=& 85.334407 + 0.0056858x_2x_5\nonumber\\
&&+ 0.00026x_1x_4 - 0.0022053x_3x_5, \nonumber\\
g_2(\bm{x}) &=&80.51249 +0.0071317x_2x_5\nonumber\\
&&+ 0.0029955x_1x_2 + 0.0021813x^2_3, \nonumber\\
g_3(\bm{x}) &=& 9.300961 +0.0047026x_3x_5\nonumber\\
&&+ 0.0012547x_1x_3 + 0.0019085x_3x_4,\nonumber\\
&&0\leq g_1(\bm{x})\leq92,\nonumber\\
&&90\leq g_2(\bm{x})\leq110,\nonumber\\
&&20\leq g_3(\bm{x})\leq25,\nonumber\\
&&78\leq x_1 \leq102,\nonumber\\
&&33\leq x_2 \leq45,\nonumber\\
&&27\leq x_3 \leq45,\nonumber\\
&&27\leq x_4 \leq45,\nonumber\\
&&27\leq x_5 \leq45,\nonumber\\
\end{eqnarray*}

The results are listed in Table \label{table2} whose corresponding experiments for the BAS-WPT algorithm just need only one beetle to run $200$ instance. Evidently, the best result generated from the BAS-WPT shows the most excellent performance among all the results listed in Table \label{table2}. The above experiments justify that the proposed BAS-WPT algorithm is effective to handle constraint optimum problem and could achieve a good performance with high convergence rate.

\begin{table*}[!t]
\footnotesize
  \centering
  \caption{Comparisons of results for Himmelblau function}
  \begin{tabular} {c c c c c c c c c c}
  \hline
&$x_1$&$x_2$&$ x_3$&$x_4$&$x_5$&$g_1({\bm{x}})$&$g_2({\bm{x}})$&$g_3({\bm{x}})$ & $f_*$\\
\hline
 \cite{meobo}   &78.00& 33.00& 29.995256& 45.00& 36.775813& 92& 98.8405& 20 &-30665.54    \\
 \cite{applied}   &78.6200 &33.4400& 31.0700 &44.1800 &35.2200 &90.5208& 98.8929 &20.1316& -30373.949\\
 \cite{gaaed}   &81.4900& 34.0900& 31.2400& 42.2000& 34.3700 &90.5225 &99.3188& 20.0604& -30183.576 \\
 \cite{covga}   &78.00& 33.00& 29.995& 45.00& 36.776& 90.7147& 98.8405& 19.9999$\diamond$&-30665.6088\\
 BAS-WPT&  78.00 &33.00 &27.1131&45.00&45.00&91.9997&100.4170&20.02056&-31011.3244\\
\hline
\end{tabular} \label{table2}
\begin{tablenotes}
 \item $\diamond$ denotes the violation of the corresponding constraint.
\end{tablenotes}
\end{table*}

\section{Conclusion}\label{sec.conclusion}

This paper extends nature-inspired BAS algorithm to solve multi-objective optimization problem and relax it to version without parameter tuning. Two typical benchmarks are considered to validate performances of the algorithm Numerical results justify the efficacy of the proposed algorithm.

%
%
%
%
%


\begin{thebibliography}{99}

\bibitem{gaiso}
D. E. Goldberg, \emph{Genetic Algorithms in Search, Optimization, and Machine
Learning,} Reading, U.K.: Addison-Wesley, 1989.
\bibitem{aapff}
 B. Tessema, and G. G. Yen, ``An adaptive penalty formulation for constrained
125 evolutionary optimization,'' \emph{IEEE Trans. Syst., Man, Cybern. A, Syst. Humans},
vol. 39, no. 3, pp. 565--578, 2009.
\bibitem{chime}
 Y. G. Woldesenbet, G. G. Yen and B. G. Tessema, ``Constraint handling in multiobjective
evolutionary optimization,'' \emph{IEEE Trans. Evol. Comput.}, vol. 13, no. 3, pp. 514--525, 2009.
\bibitem{sbice}
 T. P. Runarsson, X. Yao, ``Search biases in constrained evolutionary optimization,''
\emph{IEEE Trans. Syst., Man, Cybern. A, Syst. Humans}, vol. 35, no. 2, pp. 233--243, 2005.
\bibitem{BAS}
 X. Y. Jiang, and S. Li, ``BAS: beetle antennae search algorithm for optimization
problems,'' \emph{arXiv:1710.10724v1}.
\bibitem{aiaco}
 A. Kaveh, and S. Talatahari, ``An improved ant colony optimization for constrained
engineering design problems,'' \emph{Eng. Comput.} vol. 27, no. 1, pp. 155--182, 2010.
\bibitem{agafn}
 H. L. Li, and C. T. Chou, ``A global approach for nonlinear mixed discrete programming
in design optimization,'' \emph{Eng. Optmiz.}, vol. 22, pp. 109--122, 1994.
\bibitem{hagaw}
 C. A. Coello, and N. C. Corte, ``Hybridizing a genetic algorithm with an artificial immune system for global optimization,'' \emph{Eng. Optmiz.}, vol. 36, no. 5, pp. 607--634, 2004.
\bibitem{gofsd}
 J. F. Tsai, H. L. Li, and N. Z. Hu, ''Global optimization for signomial discrete programming problems in engineering design,'' \emph{Eng. Optmiz.} vol. 34, no. 6, pp. 613--622, 2002.
\bibitem{assmf}
 S. Akhtar, K. Tai, T. Ray, ``A socio-behavioural simulation model for engineering
design optimization,'' \emph{Eng. Optmiz.} vol. 34, no. 4, pp. 341--354, 2002.
\bibitem{aalmb}
B. K. Kannan, and S. N. Kramer, ``An augmented lagrange multiplier based
method for mixed integer discrete continuous optimization and its applications
to mechanical design,'' \emph{J. Mecha. Design, Trans. ASME}, vol. 116, pp. 318--320, 1994.
\bibitem{garod}
K. Deb, ``Geneas: a robust optimal design technique for mechanical component
design,'' in \emph{Evolutionary Algorithms in Engineering Applications,} pp. 497--514, 1997.
\bibitem{uoasp}
 C. A. Coello, ''Use of a self-adaptive penalty approach for engineering optimization problems,'' \emph{Comput. Ind.}, vol. 41, no. 2, pp. 113--127, 2000.
\bibitem{gafnm}
 S. J. Wu, P. T. Chow, ``Genetic algorithms for nonlinear mixed discrete integer optimization problems via meta-genetic parameter optimization,'' \emph{Engrg. Optim.}, vol. 24, pp. 137--159, 1995.
\bibitem{niadp}
 E. Sandgren, ``Nonlinear integer and discrete programming in mechanical design optimization,'' \emph{J. Mech. Des. ASME} vol. 112, pp. 223--229, 1990.
\bibitem{anmaf}
 K. S. Lee, Z. W. Geem, ``A new meta-heuristic algorithm for continuous engineering optimization: Harmony search theory and practice,'' \emph{Comput. Methods Appl. Mech. Engrg.}, pp. 3902--3933, 2005.
\bibitem{fastf}
 X. S. Yang, ``Firefly algorithm, stochastic test functions and design optimization,'' \emph{Int. J. Bio-Inspired Comp.}, vol. 2, no. 2, pp. 78--84, 2010.
\bibitem{meobo}
 G. G. Dimopoulos, ``Mixed-variable engineering optimization based on evolutionary
and social metaphors,'' \emph{Comput. Method. Appl. Mech. Eng.}, pp. 803--817, 2007.
\bibitem{applied}
 D. Himmelblau, \emph{Applied Nonlinear Programming}, New York: McGraw-Hill, 1972.
\bibitem{gaaed}
 A. Slipinski, H. Escalona, \emph{Genetic Algorithms and Engineering Design}, Wiley, New York, 1997.
\bibitem{covga}
 A. Homaifar, S. Lai, X. Qi, ``Constrained optimization via genetic algorithms,'' \emph{Simulation}, vol. 62, no. 4, pp. 242--253, 1994.

\end{thebibliography}
\end{document}